# FlexiTerm: A more efficient implementation of flexible multi-word term recognition

Irena Spasić

Azbookus Ltd

**Abstract.** Terms are linguistic signifiers of domain–specific concepts. Automated recognition of multi-word terms (MWT) in free text is a sequence labelling problem, which is commonly addressed using supervised machine learning methods. Their need for manual annotation of training data makes it difficult to port such methods across domains. FlexiTerm, on the other hand, is a fully unsupervised method for MWT recognition from domain-specific corpora. Originally implemented in Java as a proof of concept, it did not scale well, thus offering little practical value in the context of big data. In this paper, we describe its re-implementation in Python and compare the performance of these two implementations. The results demonstrated major improvements in terms of efficiency, which allow FlexiTerm to transition from the proof of concept to the production-grade application.

## 1. Introduction

Terms are linguistic signifiers of domain–specific concepts. From a syntactic point of view, they follow the structure of noun phrases. They can be distinguished from other noun phrases by their collocational stability. These two regularities have been exploited by various algorithms for automatic term recognition (ATR), which commonly use pattern matching rules to extract phrases of predefined syntactic structures followed by statistical analysis to measure the collocational stability of their lexical content. One such method is FlexiTerm [1, 2], which implements an unsupervised approach for the extraction of multi–word terms (MWT) from a domain–specific corpus. It has been implemented as an open-source application in Java. At 96% precision, it provided a proof of concept. It has been successfully deployed in Sentinel [3], a system for social media stream processing, but the ability of the system to cope with a high influx of text data came from basing its architecture on Advanced Message Queuing Protocol (AMQP) [4], a message-oriented middleware standard, which supports flexible management of data flow and asynchronous processing. However, as a stand-alone application, its efficiency did not allow it to be used in real-time beyond a few hundred text documents offering little practical value in the context of big data. In the meantime, Python gain popularity as a programming language especially in natural language processing (NLP) due to a number of easy-to-use libraries such as Natural Language Toolkit (NLTK) [5], spaCy [6], Gensim [7], etc. In addition to some of the Java libraries used in FlexiTerm no longer being maintained, this provided an incentive to re-implement FlexiTerm in Python while resolving its efficiency-related issues.

In this paper, we describe this re-implementation of FlexiTerm in Python. We start by describing the method in detail in order to pinpoint specific changes between the two implementations. We then validate the new implementation by evaluating precision and recall before proceeding to quantify and analyse efficiency gains.

## 2. Materials and Methods

This section describes the method behind FlexiTerm. It consists of nine functional modules, which are described and illustrated by examples in Table 1.

*Author for correspondence: Irena Spasić (azbookus@gmail.com).

**Table 1.** The FlexiTerm algorithm breakdown with examples

| Step | Module | | Operations | Examples | | | |
|---|---|---|---|---|---|---|---|
| 1 | Linguistic pre-processing | 1 | Sentence splitting | Nuclear factor of activated T-cells (NFAT) is a transcription factor which is considered to be an important regulator in early T-cell activation. | | | |
| | | 2 | Tokenisation | **Token** | **Tag** | **Lemma** | **Stem** |
| | | 3 | Tagging | Nuclear | JJ | nuclear | nuclear |
| | | | | factor | NN | factor | factor |
| | | | | of | IN | of | of |
| | | | | activated | VB | activate | activ |
| | | | | T | NN | t | t |
| | | | | cells | NN | cell | cell |
| | | 4 | Lemmatisation | ( | -LRB- | ( | ( |
| | | | | NFAT | NN | nfat | nfat |
| | | | | ) | -RRB- | ) | ) |
| | | | | is | VB | be | be |
| | | 5 | Stemming | a | DT | a | a |
| | | | | transcription | NN | transcription | transcript |
| | | | | factor | NN | factor | factor |
| | | | | ... | ... | ... | ... |
| 2 | MWT candidate extraction | 1 | Pattern matching on POS tags | Nuclear/JJ factor/NN<br>T/NN cells/NN<br>transcription/NN factor/NN<br>important/JJ regulator/NN in/IN early/JJ T/NN cell/NN activation/NN | | | |
| 3 | MWT candidate normalisation | 1 | Re-tokenisation based on string similarity (on phrases) | acetyl salicylic acid ~ acetylsalicylic acid<br>acetylsalicylic acid → **acetyl salicylic acid** | | | |
| | | 2 | Removing stopwords | Hodgkin ~~'s~~ lymphoma<br>transcription ~~of different~~ genes<br>retinoic acid receptor ~~type~~ alpha | | | |
| | | 3 | Set representation | Hodgkin 's lymphoma | | {hodgkin, lymphoma} | |
| | | | | Hodgkin lymphomas | | | |
| | | | | transcription of different genes | | {transcript, gene} | |
| | | | | gene transcription | | | |
| | | | | retinoic acid receptor type alpha | | {retino, acid, receptor, alpha} | |
| | | | | retinoic acid alpha receptor | | | |
| | | | | acetylsalicylic acid | | {acetyl, salicyl, acid} | |
| | | | | acetyl salicylic acid | | | |
| | | | | serum response | | {serum, respons} | |

| | | | | | |
|---|---|---|---|---|---|
| | | | | serums responses | |
| | | | | sera responses | |
| 4 | Acronym recognition | 1 | Acronym extraction | PKC | protein kinase C |
| | | | | | protein kinase cascades |
| | | | | | parental K562 cells |
| | | 2 | Acronym disambiguation | PKC | protein kinase C |
| 5 | Acronym integration | 1 | Adding acronyms as MWT candidates | PKC | {protein, kinas, c} |
| | | 2 | Expanding acronyms nested in other MWT candidates | <span style="color:red">PKC</span> inhibitor staurosporine | {<span style="color:red">protein, kinas, c</span>, inhibitor, staurosporin} |
| 6 | Token normalisation | 1 | String similarity (on stems) | tumor ~ tumour | {tumor, necrosi, factor, alpha} |
| | | | | | {~~tumour~~ tumor, necrosi, factor} |
| 7 | MWT classification | 1 | MWT candidate scoring | calculate C-value | |
| | | 2 | MWT selection | C-value > threshold | |
| | | 3 | Dictionary construction | **ID** | **MWT variant** |
| | | | | 384 | nfat |
| | | | | | nuclear factor of activated t cells |
| | | | | 4563 | transcription factor |
| | | | | | transcriptional factor |
| | | | | 406 | t cell activation |
| | | | | | activation of t cells |
| 8 | Text markup | 1 | Dictionary lookup | phrase matching of MWT variants | |
| | | 2 | Stand-off markup | **Document** / **Start** / **Offset** / **Label** 90368794 / 242 / 35 / 384; 90368794 / 279 / 4 / 384; 90368794 / 292 / 20 / 4563; 90368794 / 371 / 17 / 406 | |
| 9 | Visualisation | 1 | Inline markup | <mark id="384">Nuclear factor of activated T-cells</mark> ( <mark id="384">NFAT</mark> ) is a <mark id ="4563">transcription factor</mark> which is considered to be an important regulator in early <mark id="406">T-cell activation</mark> . | |
| | | 2 | Visual markup | <mark id="384" class="entity" style="background: #CCEAAB;">NFAT<span style="vertical-align: middle; margin-left: 0.5rem">384</span></mark> | |
| | | 3 | Text display | Nuclear factor of activated T-cells 384 ( NFAT 384 ) is a transcription factor 4563 which is considered to be an important regulator in early T-cell activation 406 . | |

## 2.1 Linguistic pre-processing

The process of MWT recognition from free text starts by its linguistic pre-processing, which involves text segmentation including sentence splitting followed by tokenisation. Each token is tagged by its part of speech (POS), lemmatised and then stemmed. The POS tags are utilised by the second module to identify MWT candidates, whereas the stems are utilised in the third module to normalise MWT candidates. Stemming is used to neutralise inflection and derivation. Lemmatisation is performed ahead of stemming in order to neutralise inflection of irregular words. For example, if the word 'indices' was not lemmatised first, it would be stemmed to 'indic', which would differ from the stem of its singular form 'index'. Table 1 (Step 1) illustrates this process.

## 2.2 MWT candidate extraction

Once the text is pre-processed, MWT candidates are identified by matching a pattern of POS tags. The pattern is defined by a regular expression, which can be modified as a parameter if required. By default, the following regular expression is used:

((((( NN | JJ ))* NN ) IN (((NN | JJ))* NN )) | ((( NN | JJ ))* NN POS (( NN | JJ ))* NN) | ((( NN | JJ ))+ NN )

The matching phrases are then extracted as MWT candidates. Examples given in Table 1 (Step 2) show four MWT candidates, which were extracted by matching the pattern given above.

## 2.3 MWT candidate normalisation

All MWT candidates undergo normalisation so that the main sources of term variation are neutralised. First, string similarity is used to identify equivalent phrases that vary in the way in which they were tokenised. For example, 'acetyl salicylic acid' and 'acetylsalicylic acid' differ in this respect. The high similarity of these phrases is used to hypothesise that they represent the same concept and should, therefore, have the same representation. By default, the tokenisation of the phrase with the higher number of tokens is used to re-tokenise the one with the lower number of tokens. For example, 'acetylsalicylic acid' would be effectively re-tokenised as 'acetyl salicylic acid'. Subsequently, all tokens identified as stopwords are removed from further consideration and the remaining tokens are used to create a set-based representation of MWT candidates. Table 1 (Step 3) provides examples of different MWT variants having the same representation.

## 2.4 Acronym recognition

Acronyms are normally short-form representation of MWTs, so they need to be recognised to account for the corresponding MWTs. There are two ways in which acronyms can be recognised depending on the way in which they are used. Some acronyms are explicitly defined in text, while others rely on the readers' expertise to infer their intended (or implicit) meaning. The two different types of acronym usage call for two different approaches to their recognition. FlexiTerm operates under an assumption that each corpus features predominant usage of either explicit or implicit acronyms, so this hyperparameter should be set manually to trigger the corresponding acronym recognition method. For example, explicit acronym recognition is recommended for scientific articles. Conversely, implicit acronym recognition is recommended for clinical narratives. Either method can be applied on any type of corpus, but the user should be aware that the output may vary depending on the acronym recognition method used.

With minor modifications, the Schwartz-Hearts algorithm [8] was re-purposed to support recognition of explicit acronyms. This algorithm relies on scientific conventions of acronym usage, which prescribes that the first mention of an acronym within a document needs to be accompanied with its full form by placing either within brackets. Consider the following example, which defines 'ASA' as an acronym:

> Also known as Aspirin, acetylsalicylic acid (ASA) is a commonly used drug for the treatment of
> pain and fever due to various causes.

A heuristic approach is used to locate a bracketed acronym definition and align an acronym against its full form, e.g. 'acetylsalicylic acid' = 'ASA'. Explicit acronyms are defined within a single sentence, which dramatically reduces the search space of potential full forms. One only needs to analyse the acronym's local context. Implicit acronyms, on the other hand, are not explicitly defined. This makes their recognition a more complex task as the search space of potential full forms needs to be scaled from a single sentence to the whole corpus. One way of reducing the search space is based on an assumption that an acronym's full form is actually an MWT. As MWT candidates have already been identified (see Section 2.2), they can be aligned against an acronym to identify its potential full forms. For example, the following phrases match the acronym 'ASA': 'acetylsalicylic acid', 'asymptomatic hyponatremia', 'anterolateral ischemia', 'aspiration pneumonia', 'anterior ischemia', 'acute myocardial ischemia', etc. To minimise the number of coincidental matches, only potential initialisms (e.g. 'acetyl salicylic acid') are considered.

First, potential implicit acronyms are identified on the basis of their shape, which is an abstraction over the word's orthography. Specifically, we are looking for words consisting of three or more uppercased letters only. In the following sentence, this would result in identifying 'ASA' and 'HCTZ' as potential acronyms:

> Patient was kept on her home regimen of ASA, toprol, lisinopril, HCTZ.

To identify potential full forms of 'ASA', we need to find trigrams whose initials match the given acronym. For example, one such trigram is 'acetyl salicylic acid' in the following sentence:

> The patient was given acetyl salicylic acid and nitroglycerin.

As we mentioned before, rather than considering all trigrams from the corpus, we can focus only on those trigrams that have already been identified as MWT candidates. The given trigram would indeed have been previously extracted as an MWT candidate as it represents the longest match for the regular expression provided in Section 2.2.

One problem here is that multiple matches may be found in which case sense disambiguation needs to be performed. Previously, a brute-force approach based on frequency of the potential full forms was used to select the most frequent one [2]. In this re-implementation, new functionality was added to disambiguate acronym senses. A full form that occurs in most similar contexts to those that surround a given acronym is selected. Two decisions needed to be made here, first – how to represent the context and second – how to measure similarity. The simplest way would be to collect the corresponding sentences as immediate contexts and convert them into a bag-of-words representation. However, when operating on a relatively small number of sentences, inclusion of less relevant words may introduce unnecessary noise into the representation of local contexts and skew their similarity. Verbs have been shown to demonstrate selectional preferences towards certain terms [9, 10] and as such may be good indicators of a term's sense. Hence, instead of collecting all words from the corresponding sentences, we only collected verbs. Using verbs as features, we then compared acronyms to their potential full forms using cosine similarity. We opted for this measure over the alternative ones because it represents a measurement of orientation and not magnitude [11]. Namely, the proposed verb-based feature vectors may still be sparse, in which case measures such as Euclidean distance would exhibit weak discrimination [12].

## 2.5 Acronym integration

Once acronyms have been linked to their full forms, they need to be integrated into the list of MWT candidates. This requires adding each occurrence of an acronym as an MWT candidate and normalising it using its full form. If a given full form already exists as an MWT candidate, its set representation is simply re-used for the acronym, thus resulting in both acronym and its full form having an identical representation. If the full form has not been identified previously as an MWT candidate, which may happen with explicit acronyms, then the full form is normalised following the process outlined in Section 2.3. All occurrences of the acronym as well as its full form are then added as MWT candidates using the same set representation. Looking at Table 1, we can see one such case. The long form 'nuclear factor of activated T-cells' (see Step 1) was not originally extracted as an MWT candidate (see Step 2) as the word 'activated' was tagged as a verb (see Step 1), which caused the given phrase not to match the regular expression given in Section 2.2. However, this MWT candidate was successfully recovered as the full form of acronym 'NFAT' using the Schwartz-Hearst algorithm and added to the list of MWT candidates eventually leading to both being recognised as MWTs (see Step 9).

In this manner, acronyms are added as occurrences of MWT candidates. However, we also need to normalise them to their full forms when they are nested within other MWT candidates. Using an example from Table 1 (see Step 5), we can see that the acronym 'PKC' has been linked to its full form 'protein kinase C' both of which are represented using the set {protein, kinas, c}. However, the existing representation of an MWT candidate 'PKC inhibitor staurosporine' {pkc, inhibitor, staurosporin} is not compatible with the acronym's set representation. By comparing the corresponding set representations, we cannot deduce that 'protein kinase C' is effectively nested within 'PKC inhibitor staurosporine'. This is rectified by expanding all acronyms found nested within other MWT candidates. In this case, the set representation of 'PKC inhibitor staurosporine' becomes {protein, kinas, c, inhibitor, staurosporin}.

## 2.6 Token normalisation

In this module, the normalisation of MWT candidates, which started earlier in module 2.3, is now finalised. This step involves matching similar tokens, which may vary as a result of typographical errors or different spelling. For example, 'tumour necrosis factor', whose set representation would be {tumour, necrosi, factor}, would fail to be recognised as a subpart of 'tumor necrosis factor alpha', whose set representation is {tumor, necrosi, factor, alpha} due to a different spelling of the word 'tumour'. Equivalent tokens such as 'tumour' and 'tumor' can be identified using their string similarity. Once the similarity has been established, the shortest token is used to replace all other versions. In the previous example, this would result in 'tumour necrosis factor' being represented as {tumor, necrosi, factor}, which now becomes a subset of the set representation of 'tumor necrosis factor alpha' (see Table 1, Step 6). Although this step is effectively part of MWT candidate normalisation, it is deferred until after acronyms have been recognised. The reason behind this deferral is the fact that by adding acronyms as MWT candidates and expanding them within other MWT candidates (see Section 2.5), new tokens may be introduced into the set representation of MWT candidates. Now that the list of MWT candidates is complete, their normalisation can be finalised by normalising individual elements of their set representations.

## 2.7 MWT classification

Terms are linguistic representations of domain–specific concepts [13]. However, this is not necessarily one-to-one mapping. MWTs may vary with respect to orthography (spelling, hyphenation, word segmentation, case, etc.), morphology and syntactic structure. MWTs are also frequently compressed into acronyms. It is, therefore, crucial to perform normalisation of MWT candidates prior to their statistical analysis. By aggregating equivalent MWT variants into a single normalised representative of the corresponding domain–specific concept, the statistical analysis can be performed at the latent semantic level. Hence, such analysis more closely measures termhood, which represents the degree to which a stable lexical unit is related to some domain-specific concept [14]. This in turn allows us to focus on the unithood of MWT candidates, which is defined as the degree of collocational stability.

Collocations consist of frequently co-occurring words. Absolute frequency favours shorter collocations as $f(\alpha) \geq f(\beta)$ whenever $\alpha \subseteq \beta$. Kita et al. [15] introduced a reduced cost as a function of length as a way of boosting the score of longer collocations:

$$C(\alpha) = (|\alpha| - 1) \cdot f(\alpha)$$

To further reduce the cost of $\alpha$ when $\alpha \subseteq \beta$, the cost function has been modified as follows:

$$C(\alpha) = (|\alpha| - 1)(f(\alpha) - f(\beta))$$

This can be further generalised for all $\beta$ such that $\alpha \subseteq \beta$:

$$C(\alpha) = (|\alpha| - 1)\left(f(\alpha) - \sum_{\alpha \subseteq \beta} f(\beta)\right)$$

Some of the nested occurrences of $\alpha$ can be recounted by using a weight based on the size of the set $S(\alpha) = \{\beta : \alpha \subseteq \beta\}$:

$$C(\alpha) = (|\alpha| - 1)\left(f(\alpha) - \frac{1}{|S(\alpha)|} \sum_{\beta \in S(\alpha)} f(\beta)\right)$$

This is useful when the set $S(\alpha)$ consists of a large number of infrequently occurring phrases, which themselves are not collocationally stable so their impact on the cost $C(\alpha)$ should be reduced. This approach was introduced by Frantzi and Ananiadou [16] who re-purposed this formula to extract MWTs as collocationally stable noun phrases. To improve the numerical stability of their algorithm, they further modified the cost function as follows:

$$C(\alpha) = \ln|\alpha| \cdot \left(f(\alpha) - \frac{1}{|S(\alpha)|} \sum_{\beta \in S(\alpha)} f(\beta)\right)$$

What become known as the C-value formula was originally applied to word sequences. Spasic at al. [1] applied the same formula to word sets instead. Here too we use the C-value formula to score normalised set-based representations of MWT candidates. Those MWT candidates whose score exceeds a given threshold are classified as MWTs. All MWT variants that have the same normalised set-based representative are then conflated and assigned a unique identifier in an automatically constructed dictionary of MWTs (see Table 1, Step 7 for examples). Note that an extension of FlexiTerm [17], which organised the MWTs into a hierarchy based on their semantic similarity has not yet been re-implemented in Python.

## 2.8 Text markup

The dictionary of MWTs assembled in the previous step can now be used to look them up in a corpus. Phrase matching is used to identify the corresponding occurrences in a corpus. As these matches can overlap, standoff markup is used to annotate all possible matches. For example, if one MWT is nested in another one, both will be annotated. Each annotation consists of the document identifier, the character-based position of the matching phrase and the label that represents the MWT's unique identifier. Table 1 (Step 8) provides an example of standoff markup based on the sentence given in Step 1 and an excerpt from a dictionary given in Step 7.

## 2.9 Visualisation

For the purpose of visualisation, standoff markup is converted into HTML-formatted inline markup. To avoid overlapping spans, only the longest matches are marked up. A matching phrase is enclosed by the `mark` tags and the `id` attribute instantiated by the MWT's unique identifier. Visual markup is added to highlight these phrases when rendering the corresponding HTML documents in any web browser. A random colouring scheme is applied so that different MWTs are highlighted using different colours, while different variants of the same MWT are highlighted using the same colour. In addition, the MWT's identifier is also displayed. Table 1 (Step 9) provides an example of inline markup together with its visualisation.

# 3. Implementation

Table 2 summarises the differences between two implementations of FlexiTerm. Most of them are ascribed to the availability of software libraries between the two programming languages, Java and Python. Other differences arose as the result of careful consideration of the ways in which the performance could be improved, both in terms of accuracy and efficiency.

**Table 2.** The key differences between two implementations of FlexiTerm

| Functionality | Java | Python |
|---|---|---|
| Linguistic pre-processing | Stanford CoreNLP | spaCy |
| String similarity | Levenshtein distance / Jazzy | Jaro-Winkler similarity / jellyfish |
| Implicit acronym disambiguation | frequency | verb-based cosine similarity |
| Phrase matching | Mixup | PhraseMatcher |
| Visualisation | MinorThird | displaCy |

Originally, linguistic pre-processing in FlexiTerm was performed using Stanford CoreNLP [18]. In this re-implementation, Stanford CoreNLP has been replaced by spaCy, a library of NLP tools written in Cython, a programming language designed as a superset of Python that provides C-like performance [6]. String similarity, which is used for MWT normalisation (see Table 1, Steps 3 and 6), was originally based on Levenshtein distance [19] and implemented using an open-source spell checker library called Jazzy. Searching for an appropriate library in Python was an opportune time to reconsider this metric. The main disadvantage of Levenshtein distance is that it does not take into account string length. The Jaro similarity is based on the degree to which two strings have characters in common. Formally, the Jaro similarity $J$ between two string $s_1$ and $s_2$ is defined as follows [20]:

$$J(s_1, s_2) = \frac{1}{3}\left(\frac{m}{|s_1|} + \frac{m}{|s_2|} + \frac{m-t}{m}\right)$$

Here, $m$ is the number of matching characters. The $i$-th character of $s_1$ is said to match the $j$-th character of $s_2$ if the two characters are identical and $i - d \leq j \leq i + d$, where $d = \min(|s_1|, |s_2|) / 2$. Finally, $t$ is half the number of transpositions required to align the matching characters.

The Jaro-Winkler variation scales the Jaro similarity in favour of strings that share a common prefix as follows:

$$JW(s_1, s_2) = J(s_1, s_2) + l \cdot p(1 - J(s_1, s_2))$$

where $l$ is the length of a common prefix and $p$ is the scaling factor [21]. This property proved to be useful for name matching [22]. In FlexiTerm, the calculation of this similarity was performed using jellyfish [23], a Python library for approximate and phonetic matching of strings.

In its second version, FlexiTerm incorporated acronyms as variants of MWTs [2]. However, sense disambiguation was not originally supported. This re-implementation provided an opportunity to implement this functionality. As explained in Section 2.5, the fact that verbs demonstrate sectional preferences towards certain terms [9, 10] was exploited to map each acronym to the most compatible MWT.

In the Java version, to look MWTs up in free text, they are first converted into the corresponding regular expressions, which are then matched using Mixup (My Information eXtraction and Understanding Package), a simple pattern-matching language. The Mixup output can be used by MinorThird [24], a collection of Java classes for annotating text, to visualise the results. In the Python version, MWTs are passed on to PhraseMatcher [25], a spaCy class designed to efficiently match large terminology lists. The PhraseMatcher output can be visualised using displaCy [26], a part of the core spaCy library, and the corresponding markup is returned in a format ready to be rendered and exported as HTML, which can then be displayed using any web browser.

The key functional improvements in this re-implementation include better handling of hyphens during tokenisation, more sophisticated acronym recognition including their sense disambiguation and better handling of nested acronyms. New functionality includes an option to integrate inverse document frequency into the MWT scoring, display MWT concordances for quick visualisation of their contexts and improved navigation through the output (see Figure 1). All MWTs are ranked and linked to their concordances. Each concordance line is linked to the corresponding document, which is annotated with all MWTs.

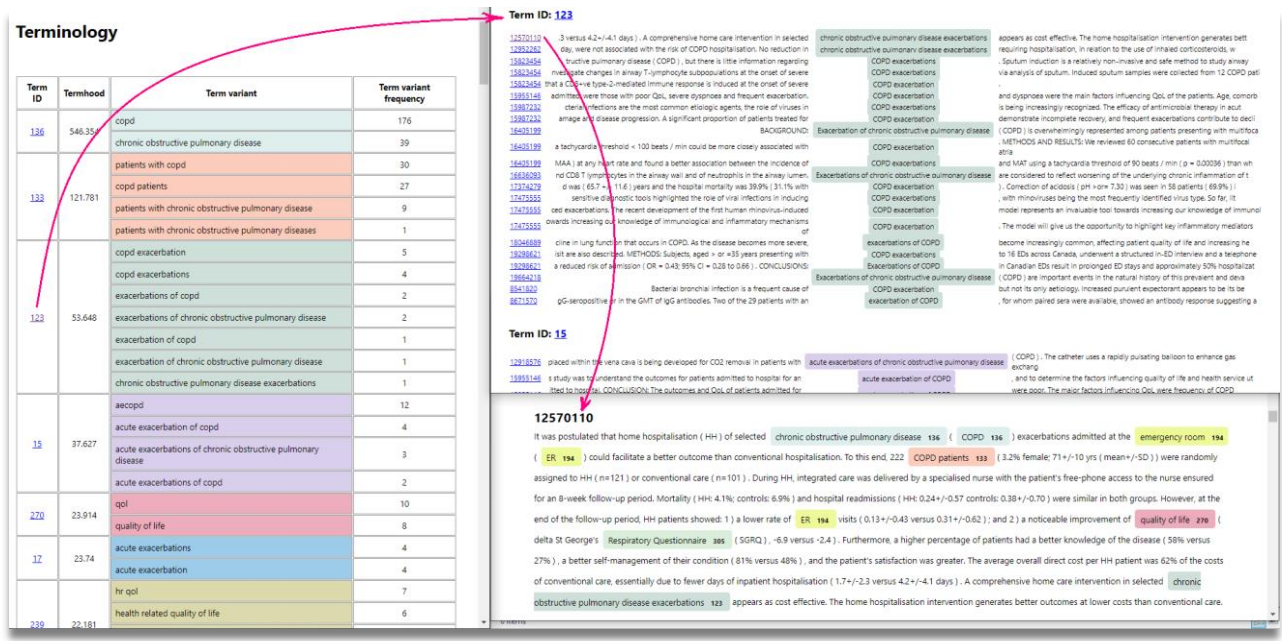

**Figure 1.** A sample output of FlexiTerm

## 4. Results

Given that the methodological changes made in this re-implementation were relatively minor, no major improvements were expected in terms of classification performance. Nonetheless, these changes warrant an evaluation to ensure that they did not introduce any unexpected errors. Therefore, we first compare the two implementations (Java vs Python) by re-using the evaluation framework originally described in [2].

Given a corpus, each implementation of FlexiTerm was applied to extract a ranked list of MWTs. Using an MWT from such list, a Boolean search query was created as a disjunction of search terms, each derived from the corresponding variant of the given MWT. For example, lets us consider MWT 'acute exacerbation of chronic bronchitis'. The corresponding Boolean query would include all of its variants recognised by FlexiTerm, e.g. 'acute exacerbation of chronic bronchitis' OR 'acute exacerbation of CB' OR AECB. The search query (represented formally in SQL) was run against individual sentences in the given corpus, which was managed in a relational database. The retrieved sentences were inspected manually to differentiate between true and false positives. A retrieved sentence was deemed a true positive if it made an explicit reference to a given MWT, otherwise it was deemed a false positive. Given the manual effort involved, this was done only for the union of $k$ top-ranked MWTs ($k$ = 20). We measured precision as the proportion of correctly retrieved sentences, i.e. P = |TP| / ( |TP| + |FP|), where TP is a set of true positives, whereas FP is a set of false positives. We measured relative recall [27] as the proportion of relevant sentences retrieved by a given FlexiTerm implementation out of all relevant sentences retrieved by either implementation, i.e. $R_i$ = | $TP_i$ | / |$TP_1 \cup TP_2$|, $i$ = 1, 2. Both precision and relative recall were micro–averaged across the 20 MWTs considered. This evaluation was repeated across five corpora consisting of 100 documents each, which are summarised in Table 3 (see [2] for more details). FlexiTerm is designed to extract domain-specific terminology. Therefore, each dataset belongs to a specific domain.

**Table 3.** Datasets used in evaluation

| Dataset | Domain | Document type | Source |
|---|---|---|---|
| D1 | molecular biology | abstract | PubMed |
| D2 | COPD | abstract | PubMed |
| D3 | COPD | patient blogpost | Web |
| D4 | obesity, diabetes | clinical narrative | i2b2 |
| D5 | knee MRI scan | clinical narrative | NHS |

Both FlexiTerm implementations achieved perfect precision meaning that all occurrences of a given search term matched the intended meaning. For example, all sentences containing an acronym 'CBF' did refer to 'core binding factor' and not another term such as 'cerebral blood flow'. Such precision can be explained by the domain-specific nature of the test corpora in which one can reasonably expect the "one sense per discourse" hypothesis to hold [28].

The relative recall values are provided in Figure 2. There were no differences in relative recall on dataset D2. Minor differences can be noticed on datasets D3 and D5. Error analysis on dataset D3 revealed a typographical error in 'pulmanary rehab', which the Python implementation failed to link to 'pulmonary rehab'. This is down to different similarity measures used and could be tackled by lowering the threshold used for the Jaro-Winkler similarity. Improved handling of hyphens led to the Python version recognising 'anti-biotics' as an MWT. Technically, this is an incorrect spelling of 'antibiotics', which is a single-word term. However, this dataset consists of patient blogposts who, as mostly laymen, may be expected to use the medical terminology incorrectly. In this case, FlexiTerm proved to be robust enough to recognise what, written like this, does appear to be an MWT. The Java version failed to tokenise the given phrase properly and consequently failed to recognise it as an MWT. Few minor differences on dataset D5 were again down to the threshold used for the Jaro-Winkler similarity being too high.

All differences on dataset D4 were related to acronyms. Even though they have all been correctly identified in both implementations, the set of acronyms that were recognised differed slightly in favour of the Python implementation. The biggest difference in relative recall can be observed on dataset D1, which was mainly related to two acronyms, 'TNF' and 'TNF-alpha', which in fact are synonyms (see [29]). The Java version does correctly link the following variants 'TNF-alpha', 'TNF' and 'tumor necrosis factor alpha', but lists 'tumor necrosis factor' separately. The Python version grouped 'TNF-alpha', 'tumor necrosis factor alpha' and a spelling variant 'tumour necrosis factor alpha' together on one hand, and on the other grouped 'TNF', 'tumor necrosis factor' and a spelling variant 'tumour necrosis factor' together. Ideally, all of these variants should be grouped as a single MWT. However, this failed to happen in either implementation because of an extra token – 'alpha'. In the context of this particular token, the Python version provides more appropriate grouping, which resulted from improved acronym disambiguation. Overall, the performance of the two implementations is comparable and, as expected due to the minor functional improvements described in Section 3, slightly in favour of the new one.

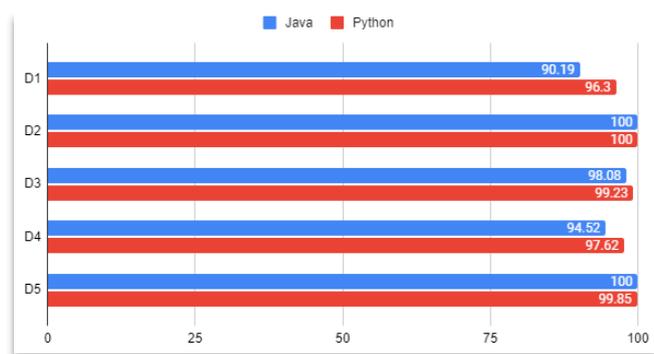

**Figure 2.** Relative recall

As mentioned earlier, the two implementations are near equivalent in terms of functionality, hence the above evaluation was performed merely as due diligence. The key performance improvements were expected in terms of scalability. To test this hypothesis, each dataset described in Table 3 was divided into 10 subsets of 10 documents each. FlexiTerm was run on the union of the first $k$ subsets ($k$ = 1, 2, … , 10) and the execution time recorded for each module. Figure 3 provides the results achieved by the two implementations. All experiments were run on a 1.6GHz Intel Core i5-8200Y CPU with 8GB RAM.

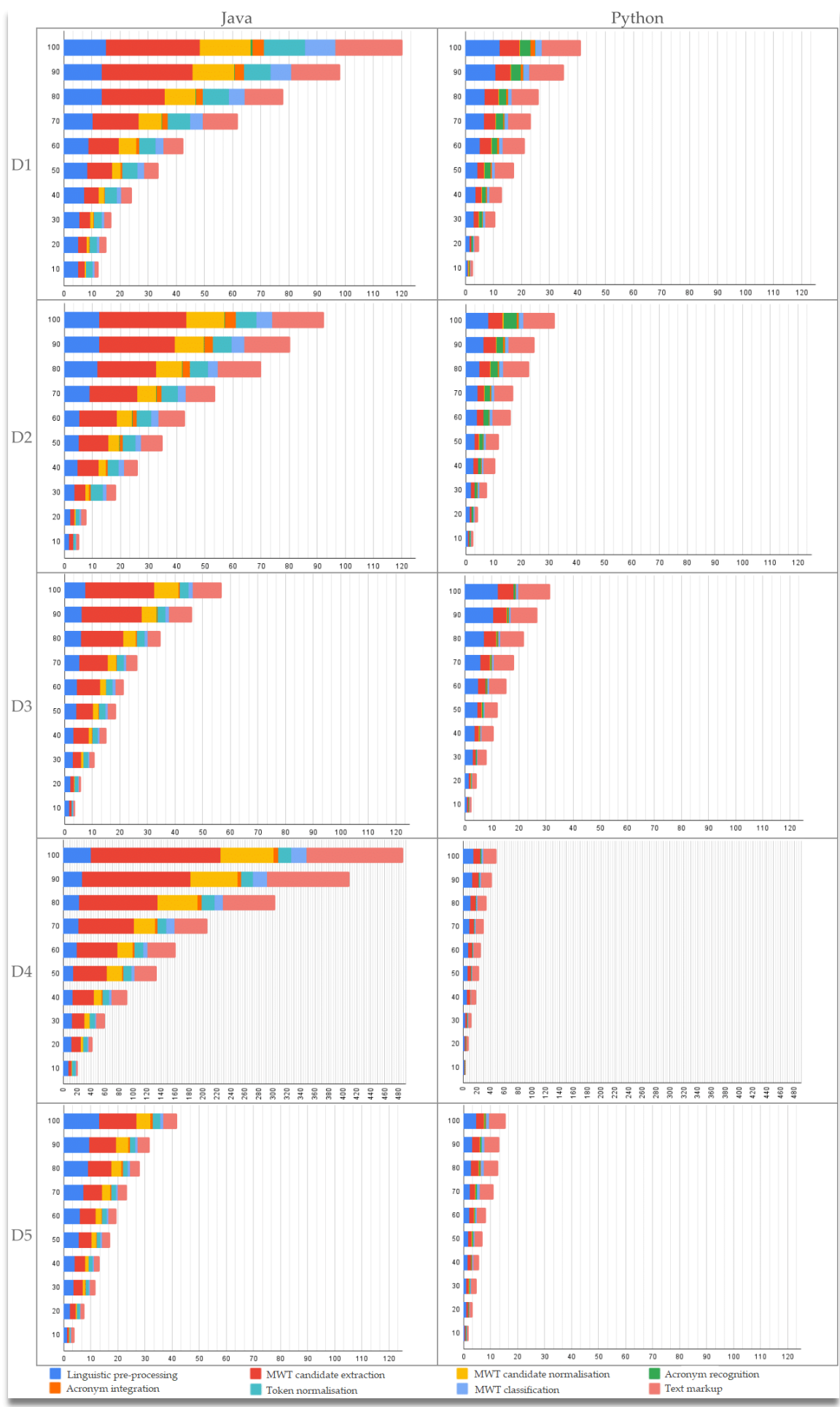

**Figure 3**. Execution times in seconds on *x*-axis for the number of documents given on *y*-axis

By contrasting the execution times across the five datasets, we can observe that linguistic pre-processing (the first step) and text markup (the last step) are fairly comparable between Java and Python. The biggest efficiency improvement can be seen in the time required to perform MWT candidate extraction. Given that no methodological changes whatsoever have been made in this module, such performance gain was unexpected given that both Java and Python use a recursive backtracking algorithm to match regular expressions despite much more efficient algorithms being available [30]. This significance of this difference is perhaps best illustrated by the fact that the Python version managed to complete all of it processing before the Java version finished extracting MWT candidates.

Other important time savings can be observed in MWT candidate normalisation and token normalisation. Both are based on string similarity, specifically Levenshtein distance [19] in the Java version and Jaro-Winkler similarity [20, 21] in the Python version. The time complexity of computing the two distances is $O(m \cdot n)$ [31, 32] and $O(m + n)$ [33] respectively, where m and n are the lengths of the two strings being compared. When a brute-force approach is applied to identify pairs of similar strings from a set of $k$ strings, a total of $k^2$ comparisons are made. The number of comparisons can be pruned substantially when the distance measure is bounded, i.e. when looking only for those pairs of strings whose similarity is greater than a given lower bound. Indeed, FlexiTerm does use a threshold, which acts as such a bound. In the case of Jaro-Winkler similarity, large differences in length are a guarantee of poor string similarity, hence a length-based filter can be used to avoid unnecessary comparisons. In addition, Jaro-Winkler similarity favours matching on prefixes, hence a prefix-based filter can be used to preselect strings that are more likely to be similar. To improve efficiency, we employed both filters. First, given a string, a prefix-based filter was used to retrieve other strings that start with the same letter. For strings that start with e (e.g. 'edema'), potential ligatures are considered by retrieving strings that start with 'ae' or 'oe' (e.g. 'oedema'). In the relational database used to manage the data in the background, all strings that need to be compared are indexed using B-trees, so that the prefix-based filter can retrieve the matching strings in logarithmic time. They are then further filtered based on their length allowing for a difference of no more than a single character. On average the normalisations steps taken together were completed in 1.5% of the total execution time using the datasets D1-D5 making these steps barely visible in the charts provided in Figure 3.

Indexing has also improved the efficiency of MWT classification, where the biggest bottleneck was identifying nested occurrences of MWT candidates as part of calculating the C-value formula. The only step that now takes longer to complete is acronym recognition. This is due to extending its functionality to include acronym sense disambiguation. The likelihood of encountering multiple matches for the same acronym increases with the corpus size. Therefore, with a view of applying FlexiTerm on larger corpora such as patent literature [34], sense disambiguation becomes an important factor in maintaining the accuracy of MWT recognition.

Overall, the performance gain increases with the amount of data to process. Looking at the results presented in Figure 3, we can see that the smallest difference can be observed on the dataset D3 where individual documents were the shortest on average. The Python version executed in 55.50% of time required for the Java version to complete its execution on a set of 100 documents. The average length of documents in the datasets D1, D2 and D5 was similar. The Python version consumed 34.36%, 34.78% and 37.40% of execution time in Java, respectively. The biggest difference was observed on the dataset D4 of lengthy discharge summaries, where only 10.00% of Java's execution time was required for Python to complete. Finally, we used the data presented in Figure 3 to estimate execution time for higher number of documents by fitting a quadratic function. The results are presented in Figure 4.

## 5. Conclusion

In this paper, we described a re-implementation of FlexiTerm, an unsupervised method for MWT recognition, in Python. The key functional improvements in this re-implementation include better handling of hyphens during tokenisation, more sophisticated acronym recognition including their sense disambiguation and better handling of nested acronyms. New functionality includes an option to integrate inverse document frequency into the MWT scoring, display MWT concordances for quick visualisation of their contexts and improved navigation through the output. Minor methodological changes were reflected in minor improvements in terms of classification performance. Major improvements in terms of efficiency now make FlexiTerm fit for practical applications as it now can complete its processing in a fraction of the previous time. The latest implementation of FlexiTerm is available under an open-source licence from https://github.com/ispasic/FlexiTerm-Python.

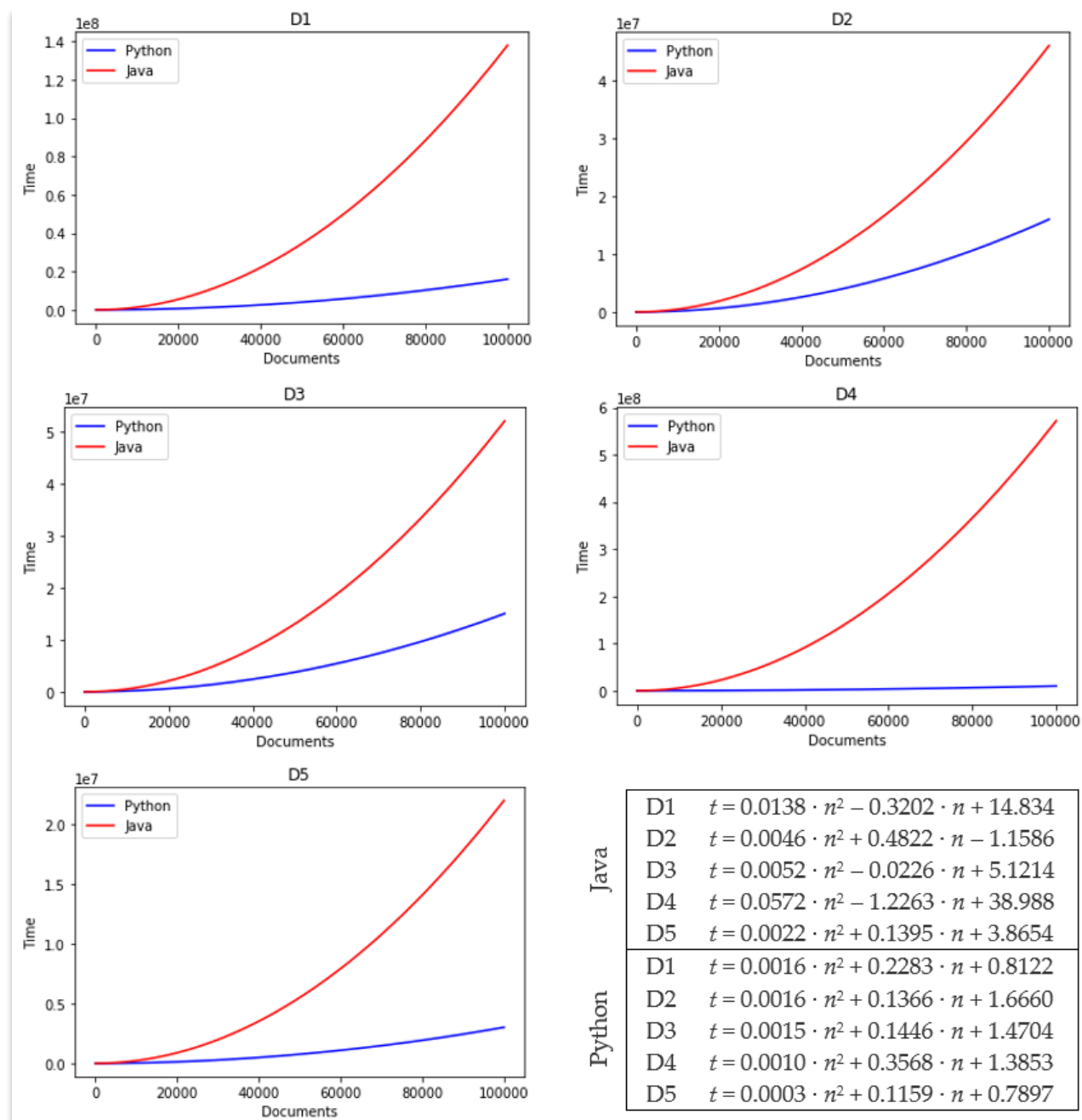

**Figure 4**. Extrapolation of execution times